\documentclass[10pt,twocolumn,letterpaper]{article}

\usepackage{iccv}
\usepackage{times}
\usepackage{epsfig}
\usepackage{graphicx}
\usepackage{amsmath}
\usepackage{amssymb}
\usepackage{flushend}
\usepackage{dirtytalk}

\iccvfinalcopy 

\def\iccvPaperID{****} 

\begin{document}

\title{Leveraging multiple datasets for deep leaf counting}

\author{Andrei Dobrescu\\
University Of Edinburgh\\
{\tt\small A.Dobrescu@ed.ac.uk}
\and
Mario Valerio Giuffrida\\
IMT Lucca\\
{\tt\small valerio.giuffrida@imtlucca.it}
\and
Sotirios A Tsaftaris\\
University Of Edinburgh\\
{\tt\small S.Tsaftaris@ed.ac.uk}
}

\maketitle
\newcommand{\mypar}[1]{\noindent\textbf{#1}}

\begin{abstract}
 The number of leaves a plant has is one of the key traits (phenotypes) describing its development and growth. Here, we propose an automated, deep learning based approach for counting leaves in model rosette plants. While state-of-the-art results on leaf counting with deep learning methods have recently been reported, they obtain the count as a result of leaf segmentation and thus require per-leaf (instance) segmentation to train the models (a rather strong annotation). Instead, our method treats leaf counting as a direct regression problem and thus only requires as annotation the total leaf count per plant. We argue that combining different datasets when training a deep neural network is beneficial and improves the results of the proposed approach. We evaluate our method on the CVPPP 2017 Leaf Counting Challenge dataset, which contains images of Arabidopsis and tobacco plants. Experimental results show that the proposed method significantly outperforms the winner of the previous CVPPP challenge, improving the results by a minimum of ~50\% on each of the test datasets, and can achieve this performance without knowing the experimental origin of the data (i.e. \say{in the wild} setting of the challenge). We also compare the counting accuracy of our model with that of per leaf segmentation algorithms, achieving a 20\% decrease in mean absolute difference in count ($|$DiC$|$).
\end{abstract}

\section{Introduction}
Plant phenotyping is a growing field that biologists have identified as a key sector for increasing plant productivity and resistance, necessary to keep up with the expanding global demand for food. Computer vision and machine learning are important tools to help loosen the bottleneck in phenotyping formed by the proliferation of data generating systems without all the necessary image analysis tools \cite{Minervini2015}. 

The number of leaves of a plant is considered one of the key phenotypic metrics related to its development and growth stages \cite{Telfer1997,Walter1999}, flowering time \cite{Koornneef1995} and yield potential. Automated leaf counting based on imaging is a difficult task. Leaves vary in shape and scale, they can be difficult to distinguish and  are often occluded. Moreover, a plant is a dynamic object with leaves shifting, rotating and growing between frames which can be challenging to computer vision counting approaches \cite{Minervini2015}.

From a machine learning perspective, counting the number of leaves can be addressed in two different ways: (i) obtaining a per-leaf segmentation, which automatically leads to the number of leaves in a rosette \cite{Ren2016,Romera-Paredes2016,Scharr2016}; or (ii) learning a direct image-to-count regressor model \cite{Giuffrida2015,Pape2015}. Deep learning approaches in this field show impressive results in obtaining leaf count as a result of per-leaf segmentation, but they require individual leaf annotations as training data, which are difficult and laborious to produce. In fact, regression approaches leverage this issue by using the total leaf count in plants as its only supervision information \cite{Ren2016,Romera-Paredes2016}. There are few annotated datasets for rosette plants \cite{Bell2016,Cruz2016,Minervini2016} which is a limitation when trying to implement deep learning approaches for plant phenotyping problems \cite{Tsaftaris2016} or for field of phenotyping in general \cite{Lobet2017}.

\begin{figure*}[t]
\includegraphics[width=\textwidth]{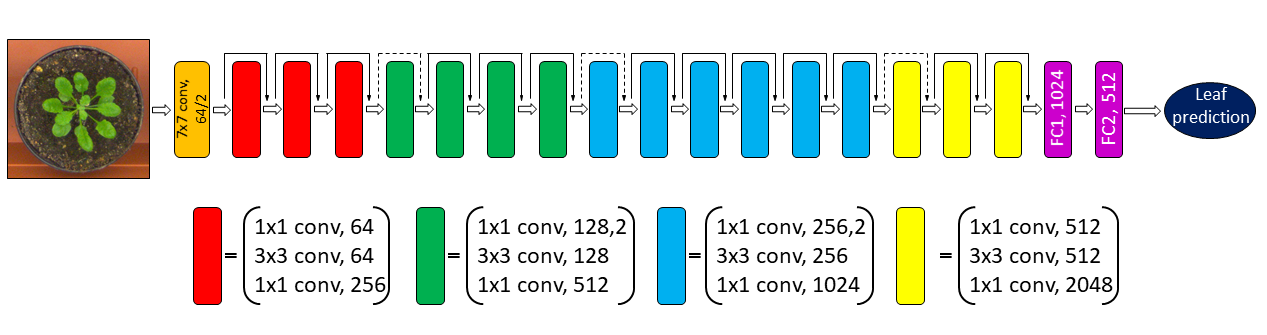}
   \caption{Architecture of our modified \textit{Resnet50} model \cite{He2015}. The network takes as input a RGB image of a rosette plant and outputs a leaf count prediction. The classification layer was removed and replaced with two fully connected layers FC1 and FC2. The network is comprised of 16 residual blocks which consist of three stacked layers each with skip connections between the input and the output of each block. The solid lines represent connections which maintain dimensions while dotted lines increase dimensions.}
\label{fig:short}
\end{figure*}

In this paper, we propose a deep learning model for leaf counting in rosette plants on top-down view images.  The backbone of the model is a modified \textit{Resnet50} deep residual network \cite{He2015} pre-trained on the ImageNet dataset (c.f. \figurename~\ref{fig:short}). The network is fine-tuned on one or more datasets and provides as output a leaf count. To boost deep learning performance to learn despite being provided with small datasets, we found that pooling data from different sources and even different species (and cultivars) for the purposes of training improves leaf prediction accuracy. Our method treats leaf counting as a direct regression problem, therefore it only requires the total leaf count of each image as annotation. We evaluate our approach on datasets provided in the \textit{Leaf Counting Challenge} (LCC) held as part of the \textit{Computer Vision Problems in Plant Phenotyping} (CVPPP 2017) workshop. The datasets consist of top-down images of single plants of Arabidopsis (A1, A2, A4) and tobacco (A3) plants collected from a variety of imaging setups and labs. In this challenge, there was also a \say{wild} test dataset (A5) which combines test images from all the other datasets in order to assess the generalization capabilities of machine learning algorithms without knowing the experimental origin of the image data. The final model is robust to nuisance variability (i.e. different backgrounds, soil) and variations in plant appearance (i.e. mutants with altered plant shape and scale). We employed an ensemble method of four models to obtain the results of the LCC challenge.  Experimental results show that our approach outperforms the winner of the previous CVPPP challenge \cite{Giuffrida2015} as well as a state-of-the-art counting via segmentation approach.

The remainder of this paper is organized as follows. In Section \ref{sec:rel} we review the current literature. In Section \ref{sec:meth} we detail our deep learning network. Then, in Section  \ref{sec:trainingset} we report the experimental results. Particularly, the results of the CVPPP challenge on the testing set are reported in Section \ref{sec:challenge}. Finally, Section \ref{sec:conclusions} concludes the paper.


\section{Related Work}
\label{sec:rel}


\mypar{Counting via detection.} One class of approaches involves counting by detection \cite{Yao2012} which frames the problem as a detection task. Some solutions rely on local features such as histogram orientated gradients \cite{Chayeb2014} or shape \cite{Dalal2005}. Object detectors using region based convolutional neural networks \cite{Girshick_2015_ICCV} have attracted attention by providing state-of-the-art detection results while reducing training and testing times. They incorporate region proposal \cite{girshick2014rcnn,ren2015faster} and spatial pyramid pooling networks \cite{He2014} to provide region of interest suggestions and then fine-tuning the resulting bounding boxes to fit on the desired objects. 

\mypar{Counting via object segmentation.} Object detection is considered an easier task than segmentation. In fact, once an object is detected in the scene, obtaining a per-pixel segmentation mask is not trivial. Especially for the case of multi-instance segmentation \cite{Romera-Paredes2016}, where the same objects appear multiple times in an image (e.g., leaves of a plant). Pape and Klukas \cite{Pape2015} used split points to determine lines between overlapping leaves to assign them different labels. Deep learning solutions using recurrent neural networks achieve state-of-the-art leaf segmentation and counting results. In \cite{Romera-Paredes2016}, the authors developed an end-to-end model of recurrent instance segmentation by combining convolutional LSTM \cite{Donahue_2015_CVPR} and spatial inhibition modules as a way to keep track of spatial information within each image allowing to segment one leaf at a time. The method also deploys a loss function which learns to segment all separate instances sequentially and allows the model to learn and decide the order of segmentation. In \cite{Ren2016}, another neural network that uses visual attention to compute instance segmentation jointly with counting was proposed. This method has sequential attention by creating a temporal chain via a LSTM cell which outputs one instance at a time. Non-maximal suppression, used to solve heavily occluded scenes, is dynamically leveraged using previously segmented instances to aid in the discovery of future instances. 

\mypar{Counting via density estimation.} Another method to count objects in an image is by estimating their distribution, using local features.  In \cite{Lempitsky2010}, the authors have developed a loss function which aims to minimize Maximum Excess over SubArrays (MESA) distance. Other methods include density estimation by per-pixel ridge and random forest regression. Similar approaches can be found in \cite{Arteta2014,Arteta2016,Fiaschi2012}, where regressors are used to infer local densities. However, these approaches are difficult to use for leaf counting, as they are challenged by the huge scale variability of leaves, as well as heavy occlusions and overlaps.

\mypar{Direct count.} Leaf counting results using machine learning solutions have been reported in past CVPPP challenges as well as in other independent reports which have identified plant datasets as compelling ways to test models. The winner of the previous CVPPP challenge \cite{Giuffrida2015} adopted a direct regression model through support vector regression. The method involved converting the image into a log-polar coordinate system before learning a dictionary of image features in an unsupervised fashion. The features were learned only in regions of interest determined by texture heuristics. The use of the log-polar domain provided the method with rotation and scale invariance, however the scale of the leaves is an important feature to learn, as is can vary considerably within a plant and is directly correlated with the growth stage of the plant. In \cite{Pape2015}, the authors used a set of geometrical features to fit several classification and regression models. Using different tools available in WEKA \cite{Hall2009}, they found that the Random Subspace method \cite{Ho1998} could obtain lowest $|$DiC$|$ only using geometrical features.

\begin{figure}[t]
\centering
\includegraphics[width=\linewidth]{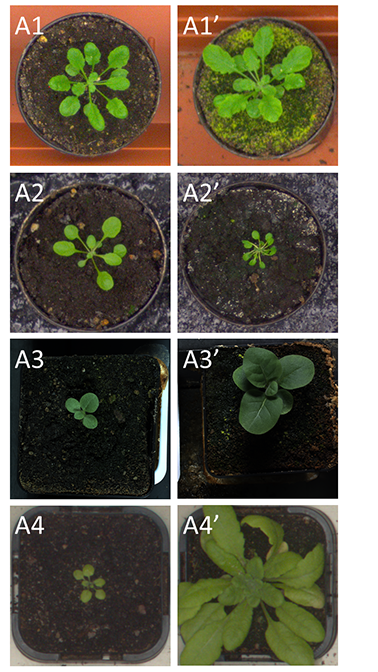}
   \caption{Examples of rosette plant images present in the four training datasets A1, A2, A3, and A4. The datasets show the big variety between images for applications in plant phenotyping. The left column represents images which have more well defined and easier to identify leaves, making leaf count relatively easier to determine. The right column shows examples of more challenging images for computer vision applications due to difficult to distinguish backgrounds (A1'), mutants which alter plant appearance (A2') and heavily occluded leaves (A3', A4').}
\label{fig:dataset}
\end{figure}

\section{Methodology}
\label{sec:meth}

We implemented a deep learning approach for counting leaves in rosette plants. We used a modified \textit{ResNet50} \cite{He2015} residual neural network to learn a leaf counter taking as input a top-down view RGB image of a rosette plant. For this paper, we have adhered to the guidelines and data provided for the CVPPP leaf counting challenge. 

\subsection{The Network}
The architecture of our model is displayed in \figurename~\ref{fig:short}. We used a \textit{Resnet50} because of its ability to generalize, which was crucial for this challenge for its \say{in the wild} setting, as well for its fast training and convergence speed. The ResNet architecture is easier to optimize than other deep networks and addresses the degradation problem present in very deep networks which states that as deep networks converge and accuracy gets saturated, it starts to degrade \cite{He2015}. The problem is addressed by the residual convolutional blocks which make up the network which is described as follows. If $H(x)$ is an underlying mapping of several stacked layers and we assume that the layers can approximate a complex function then we can assume that they can also approximate the residual function $F(x) = H(x) - x$. This changes the original function from $H(x)$ to $F(x)+x$. Furthermore, identity mappings, implemented as skip connections, ease optimization because they help propagate the error gradient signal faster across all layers. The positive impact on optimization and learning grows with increased layer depth. \cite{HeIdentity}. The residual functions are learned with reference to the layer inputs facilitated by the skip connections between the residual blocks as seen in \figurename~\ref{fig:short}.

We modified the reference \textit{ResNet50} by removing the last layer intended for classification, flattening the network and adding two fully connected layers FC1 and FC2 of 1024 and 512 nodes respectively followed by ReLU activations. We apply an L2 activation regularization on the FC2 layer to penalize the layer activity during training and prevent overfitting. FC2 goes into a fully connected layer containing a single node which acts as the leaf count prediction.

\subsection{The datasets}
The challenge consisted of four RGB image training datasets noted as A1, A2, A3 and A4 \cite{Bell2016,Minervini2016}. 
A1 and A4 contain images of wild-type Arabidopsis plants (Col-0) while A2 contains four different mutant lines of Arabidopsis which alter the size and shape of leaves. A3 is made up of young tobacco plant images. The training sets include 128, 31, 27, 624 images and the testing sets contain 33, 9, 65, 168 images for A1, A2, A3, A4 respectively. The datasets are taken from different labs, with different experimental setups, and thus background appearance and genotype composition varies (\figurename~\ref{fig:dataset}). Furthermore, the images in each dataset have different dimensions ranging from 500x530 pixels in A1 to 2448x2048 pixels in A3. For testing, in addition to the four aforementioned datasets, the organisers provided a \say{wild} dataset A5 which combines images from all testing datasets, to test methods that generalize across data and which are not fine-tuned (and specific) for each dataset. For training, we also formed a dataset similar to the \say{wild} dataset, named Ac, in which we combined all the images in the four training datasets.

\subsection{Training procedure}
For pre-processing, each image was re-sized to be 320x320x3 pixels and a histogram stretch was applied on all images to improve contrast as some images were darker than others. The resolution was chosen to optimize training times while retaining important features such as distinguishable small leaves. We used a random split from the training set to $50\%$ of the images used for training $25\%$ for validation and $25\%$ for (internal) testing. The split ratio was chosen so that there would be a similar percentage of test images per run as the test set of the challenge. Furthermore, the training, validation and test sets include plants of all ages by taking an even distribution from each dataset.  To evaluate our architecture, we first trained the network on each of the CVPPP datasets individually. We then added more data by combining datasets together (i.e. A1 + A2) and finally a combination of all four datasets named Ac. The test results from the combinations we evaluated can be seen in Table \ref{tab:table1}. Cross validation was performed four times on differently sampled training images when training on (Ac). We used mean squared error (MSE) as the loss function and Adam optimizer \cite{Kingma2015} with a learning rate of 0.001. We trained with an early stopping criterion, based on the validation loss to avoid overfitting.

Data augmentation was performed when training all models. We used a generator which assigns training images a random affine transformation from a pool of random rotation from 0-170 degrees, zoom between $0-10\%$ of the total image size and flipping vertically or horizontally. The training steps for each epoch was defined as the double the number of training images and the batch size per step was 6. In total, the augmented dataset was 12 times the size of the original set per training epoch. Once the models are trained, obtaining test predictions just requires inputting the desired test images. The network output is not discrete, so we round the predictions to the nearest integer to get a leaf count.

For the challenge test set, we employed an ensemble method comprised of four models trained on four different equal portions of the Ac dataset. We fused the predictions of the four models by averaging them to obtain the results shown in Table \ref{tab:lccres}.

\subsection{Implementation details}
We implement our models in Python 3.5. For training, we used an Nvidia Titan X 12Gb GPU using Tensorflow as backend. The models took between 1.5 and 5 hours to train depending on how many datasets were pooled together, over an average of 50 epochs.

\subsection{Evaluation metrics}
We used the same evaluation metrics as those provided by the organizers of the workshop to assess our network’s performance: Difference in Count (DiC), absolute Difference in Count  $\vert$DiC$\vert$, mean squared error (MSE) and percent agreement given by the percentage of exact predictions over total predictions. For our internal testing, we also include the R2 coefficient of determination.

\begin{table} [t]
\centering
{\small
\begin{tabular}{cccccc}
    \multicolumn{6}{c}{(A) Test results on training set A1}                                      \\ \hline
Train Sets & DiC         & $|$DiC$|$      & MSE  & R$^2$ & {[}$\%${]}  \\ \hline
A1            & -0.81(0.85) & 0.94(0.70) & 1.38 & 0.23                & 25                 \\
A1+A2         & -0.06(1.03) & 0.75(0.71) & 1.06 & 0.76                & 41                 \\
A1+A4         & -0.75(0.90) & 0.88(0.78) & 1.38 & 0.69                & 34                 \\
Ac            & 0.28(0.80)  & 0.53(0.66) & 0.72 & 0.60                & 56     \\ \hline           
\end{tabular}
}

\end{table}

\begin{table}[t]
\centering
{\small
\begin{tabular}{cccccc}
    \multicolumn{6}{c}{(B) Test results on training set A2}                                  \\ \hline
Train Sets & DiC         & $|$DiC$|$      & MSE  & R$^2$ & {[}$\%${]}  \\ \hline
A2         & -2.38(2.69) & 2.38(2.69) & 12.88 & 0.29                & 38        \\
A1+A2      & -0.56(2.06) & 1.69(1.51) & 6.31  & 0.65                & 25        \\
A2+A4      & -0.75(2.15) & 1.75(1.45) & 6.38  & 0.65                & 31        \\
Ac         & -0.38(1.11) & 0.88(0.78) & 1.38  & 0.92                & 38    \\ \hline   
\end{tabular}
}
\end{table}

\begin{table}[t]
\centering
{\small
\begin{tabular}{cccccc}
\multicolumn{6}{c}{(C) Test results on training set A3}                               \\ \hline
Train Sets & DiC         & $|$DiC$|$      & MSE  & R$^2$ & {[}$\%${]}  \\ \hline
A3         & -0.57(1.50) & 1.43(0.73) & 2.57 & 0.46                & 14      \\
Ac         & 0.71(1.03)  & 0.71(1.03) & 1.57 & 0.47                & 57   \\ \hline  
\end{tabular}
}

\end{table}

\begin{table}[t]
\centering
{\small
\begin{tabular}{cccccc}
         \multicolumn{6}{c}{(D) Test results on training set A4}                                 \\ \hline
Train Sets & DiC         & $|$DiC$|$      & MSE  & R$^2$ & {[}$\%${]}  \\ \hline
A4         & 0.1(1.14)   & 0.91(0.85) & 1.54 & 0.96                & 35      \\
A1+A4      & -0.01(1.06) & 0.77(0.73) & 1.12 & 0.97                & 39      \\
A2+A4      & 0.05(1.04)  & 0.73(0.75) & 1.10 & 0.97                & 43      \\
Ac         & 0.12(0.99)  & 0.69(0.73) & 1.01 & 0.97                & 46 \\ \hline    
\end{tabular}
}
\caption{Evaluation results of models tested on just the training datasets. The first column of each table represents training regimen of our network comprising of single and combined datasets (Ac denotes a combination of all datasets). The values were obtained through cross-validation using a split of 50\%, 25\%, 25\% images for training, validation and testing respectively. Test sets all refer to our internal split of the original training set as described in text.}
\label{tab:table1}
\end{table}

\begin{figure*}[th]
\begin{center}
\includegraphics[width=\textwidth]{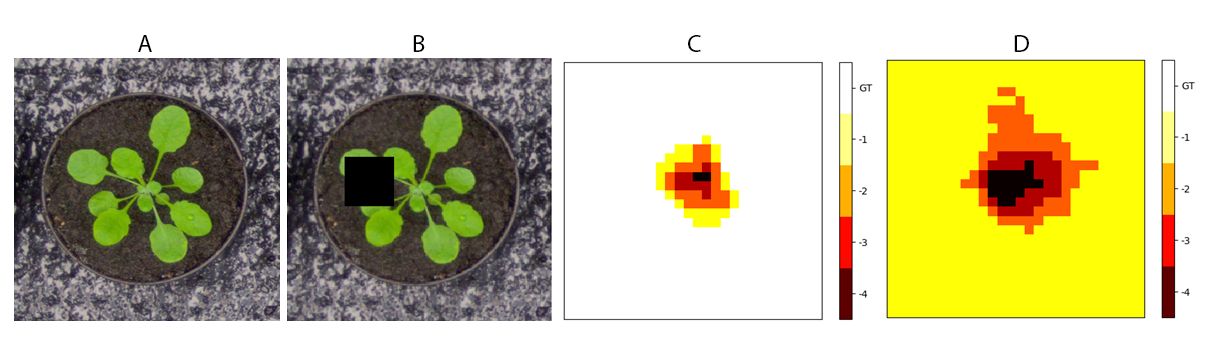}
\end{center}
   \caption{Test showing network training ability by obscuring part of the image with a sliding window. (A) is the original image and (B) shows the black sliding window (60x60 pixels) traversing the original image. (C) represents a heatmap of the accuracy of leaf count prediction as the sliding window is traversing the image using a model trained on the Ac dataset, showing that the errors are confined only to the area where the plant is located. (D) is the prediction heatmap on a model trained on just the A2 dataset (the origin of the image).}
\label{fig:train}
\end{figure*}

\section{Results on the training set and discussion}
\label{sec:trainingset}
We first tested our architecture on the training datasets in order to assess its performance and the results can be seen in Table \ref{tab:table1}. The tests were performed following the same training regimen outlined in the Section \ref{sec:meth}. We found that fine tuning the \textit{Resnet50} network, pretrained on the ImageNet dataset, gave better and more consistent results than providing stronger  annotations and using random initialization. So the learned ImageNet features were more valuable for this task than having the segmentation mask as an input.

Our solution was able to overcome several challenges. Firstly, the network can work with images of different sizes and scales present in each dataset. That said it is important that the original quality of the images is good enough to distinguish between leaves. Secondly, our model was able to learn to count leaves of different shapes, sizes and orientation provided only with minimal annotations. As seen in \figurename~\ref{fig:dataset}, the plants were of a diverse nature in terms of age, genotype and species giving a wide degree of complexity in counting. This ties in with one of the limitations of the direct regression approaches, namely that the network has to infer more information from each image to compensate for the lack of stronger annotations. This can be attenuated by providing more data when training to give a better chance of learning relevant features. Labeling data is increasingly time and resource intensive when going from weak to stronger annotations (e.g total leaf count vs. per leaf segmentation mask) which is one of the reasons for the relative lack of publicly available plant phenotyoing datasets. By employing models which require only weak annotations for training as opposed to models which require strong annotations \cite{Ren2016}, it becomes possible to have access to more labelled data given limited resources. Lastly, the provided datasets contained a limited amount of training images compared to the quantities of data traditionally employed for deep learning models. Furthermore, they were not uniformly represented with set A2 including just 31 images, while set A4 had 623. Even with data augmentation, there was a barrier to how much the network would learn when trained on each dataset separately, as evident in Table~\ref{tab:table1}. We sought to improve those initial results by combining the datasets together when training, so to provide more varied and meaningful examples than the ones supplied solely by data augmentation. 

\begin{table}
\begin{center}
{\small
\begin{tabular}{cccccc}
\hline
       & DiC   & $|$DiC$|$ & MSE & R$^2$ & {[}$\%${]}  \\  \hline
A1*    & -5.46(0.83) & 5.46(0.82)  & 30.59     & 0.07                      & 2         \\
A1     & -1.17(0.57) & 1.17(0.57)  & 1.70      & 0.38                      & 8         \\
A1+A4* & -0.58(0.59) & 0.59(0.59)  & 0.69      & 0.98                      & 46 \\ \hline      
\end{tabular}
}\end{center}
\caption{Internal test metrics from models with and without (*) augmentation. The sets indicated with a * in the name (e.g. A1*) were trained without using data augmentation.}
\label{tab:tab2}
\end{table}

\begin{table*}[t]
\begin{center}
{\small
\begin{tabular}{l|ccc|cccc|cc|cc}
\hline
 & \multicolumn{3}{c}{DiC} & \multicolumn{4}{c}{$|$DiC$|$} & \multicolumn{2}{c}{Agreement {[}\%{]}} & \multicolumn{2}{c}{MSE} \\ \hline
 & Ours & Ref\cite{Giuffrida2015} & Ref\cite{Romera-Paredes2016} & Ours & Ref\cite{Giuffrida2015} & Ref\cite{Romera-Paredes2016} & Ref\cite{Ren2016} & Ours & Ref\cite{Giuffrida2015} & Ours & Ref\cite{Giuffrida2015} \\ \hline
A1 & -0.39(1.17) & -0.79(1.54) & 0.2(1.4) & 0.88(0.86) & 1.27(1.15) & 1.1(0.9) & 0.8(1.0) & 33.3 & 27.3 & 1.48 & 2.91 \\ 
A2 & -0.78(1.64) & -2.44(2.88) & - & 1.44(1.01) & 2.44(2.88) & - & - & 11.1 & 44.4 & 3.00 & 13.33 \\ 
A3 & 0.13(1.55) & -0.04(1.93) & -  & 1.09(1.10) & 1.36(1.37) & - & - & 30.4 & 19.6 & 2.38 & 3.68 \\ 
A4 & 0.29(1.10) & - & - & 0.84(0.76) & -  & -  & -  & 34.5 & - & 1.28 & - \\ 
A5 & 0.25(1.21) & - & - & 0.90(0.85) & - & - & - & 33.2 & - & 1.53 & - \\ 
All & 0.19(1.24) & - & - & 0.91(0.86) & - & -  & - & 32.9 & - & 1.56 & - \\ \hline
\end{tabular}
}
\end{center}
\caption{Results for the test set for the provided datasets based on the fused ensemble of models trained on dataset Ac. For comparison we include the results of the winner of the previous CVPPP leaf counting challenge \cite{Giuffrida2015} as well as two count derived from per leaf segmentation approaches \cite{Romera-Paredes2016,Ren2016}.}
\label{tab:lccres}
\end{table*}

Results show that combining datasets from different sources and even species is beneficial, since it improves test accuracy for all datasets and more generally for all evaluation metrics (Table 1). The models were trained using data augmentation procedures mentioned above. The worst performance is seen in networks which are trained only on a single dataset. The combination of any two datasets yields similar results, no matter which combination is used even though A4 is a larger dataset than A2 and A1. The best results are given by the models trained on the grouping of all datasets (Ac). The degree of the improvement varies, with the most accentuated being A2 and the least impacted being A4. That was not unexpected, as A2 is the smallest dataset so additional data would cause a large impact in learning while A4 is by far the largest dataset. However, even in A4, the MSE and mean absolute difference in count ($|$DiC$|$) decreases by 30\% when training on combined datasets compared to just on A4 (Table \ref{tab:table1}D).

There was also a significant improvement between models trained on just A3 compared to Ac (Table \ref{tab:table1}C), especially in terms of $|$DiC$|$ reducing it by 50\%, MSE by 40\% and percent agreement improved 4-fold. As the models trained on just A3 and Ac were presented with the same number of tobacco plant images, we see that the Ac model shows improved learning even when the only extra data comes from other plant species.

One natural question that arises is whether the network is learning to actually look at leaves to count or is influenced by the material in the background (i.e. it relies on background cues).  We tested the network's ability to learn by imposing a black sliding window on images used in training \cite{Zeiler2014}. We predicted the leaf count using our models on the images as the sliding window was traversing the image to see if by obstructing the portion of the image the network was meant to learn it would give rise to errors in count. By sliding across the image and reporting a result (of prediction error) per position of the sliding window, a heat map of errors is generated. An example of this process is shown in \figurename~\ref{fig:train}. We used the sliding scale test on models trained on the aggregated dataset Ac (\figurename~\ref{fig:train}C) and on models trained on just the dataset which contained the original image (\figurename~\ref{fig:train}D). The errors in the model trained on dataset Ac were very specific to regions containing the plant suggesting that the model learned well the area of interest. As the sliding window moved closer to the center of the plant the errors increase as the window obstructs smaller leaves. The model trained on only one dataset did not perform as well, having a 1 difference error in general (yellow homogeneous background) and the regions which were affected by the sliding box were not only specific to the plants location.

We believe that the improved results are a consequence of increased data variability which allows the model to learn more precise features. To rule out that improved results could also be reached just by additional image augmentation we tested a combined trained model of A1+A4 with no augmentation versus a model of just A1 with data augmentation that would bring the total number of training images to be equal to that of A1+A4. The dataset A1 was chosen for this experiment because it is the most balanced dataset in terms of size.  The results show a decreased test MSE of by more than 100\% and the R\textsuperscript{2} coefficient went from 0.38 to 0.98 (Table \ref{tab:tab2}) in the models trained on the three datasets without augmentation compared to the single A1 with augmentation.

\section{Results on the testing set as provided by the challenge organizers and discussion}
\label{sec:challenge}
The results of the LCC were compiled by the organizers using the metrics provided (Table \ref{tab:lccres}). We did not use the provided segmentation masks or the leaf center dots in our models. We used an ensemble method comprised of four models of the same architecture but trained on different equal portions of the Ac combined dataset. We ran each of the models on the test set and fused the leaf count predictions by averaging the outputs to reach the results in Table \ref{tab:lccres}. The A3 image dataset is an important indicator of how well a model learned to generalize because it contains 27 images in the training set and 65 images in the testing set while having the only tobacco images in the challenge. 

We outperformed the winner of the previous CVPPP LCC by at least 50\% for each of the datasets provided in MSE and $|$DiC$|$ without needing to know the experimental origin of the data (i.e. \say{in the wild} setting of the challenge). We achieved an average of 33\% agreement across the datasets with the exception of A2 where we had lower percent agreement than than the reference although our approach resulted in a significantly lower MSE. Training our models on the combined set Ac helped us achieve similar or better results for the A5 \say{wild} dataset compared to that of individual datasets. 

The previous challenge only provided datasets A1, A2 and A3 to the participants, thus we can only compare results on those three datasets. Overall, we had an improvement on all parameters for all three datasets, with the largest improvements in A2, which represented the smallest Arabidopsis dataset and contained difficult mutants with altered shape and size.

We also compared our results with the state-of-the-art per leaf segmentation approaches from \cite{Romera-Paredes2016} and \cite{Ren2016}. We outperformed \cite{Romera-Paredes2016} by achieving a 20\% reduction in mean $|$DiC$|$ compared to them. We also obtained similar results to \cite{Ren2016}, with the advantage of requiring less strong annotations compared to their method.

\section{Conclusions}
\label{sec:conclusions}

Here we presented a deep learning approach for leaf counting in rosette plants from top-down view RGB images. We showed that it is possible to use deep learning as a way of directly performing automated leaf counting for plant phenotyping and improve upon the current state of the art.
Firstly, we implemented a modified ResNet50 deep residual neural network to act as a leaf prediction model where we treated leaf counting as a direct regression problem using only the total leaf count per plant as required annotation. Secondly, we found that pooling data from different sources for the purposes of training improves leaf prediction accuracy. Importantly, we also discovered that training on aggregated datasets also provides the model invariance to plant species, which is an important factor when testing the model with the \say{wild} set.

We evaluated our network on the standardized datasets provided in the context of the Leaf counting challenge of the CVPPP 2017 workshop and then compared to previously published networks. We found that our method outperforms the previous winner of the challenge by at least 50\% for all the provided datasets. Furthermore, we compared our results with count from per leaf segmentation approaches and we achieved improved and comparable results to the two state-of-the art solutions currently available. In the first case, we outperformed the state-of-the-art network by decreasing the  mean $|$DiC$|$, and in the second case we required far less strong annotation to achieve comparable results.

For the purposes of this challenge we only had access to images provided, so we could not investigate or optimize other parameters such as ideal camera placement for this task, but it would be an interesting avenue to pursue. 

\section{Acknowledgements}
\label{sec:acknowledge}
This work was supported by an EPSRC DTP PhD fellowship, number EP/N509644/. We acknowledge and thank NVIDIA for providing hardware essential for our work. Finally we would like to thank all the organisers of the CVPPP workshop for making it happen.

{\small
\bibliographystyle{ieee}
\bibliography{egbib}
}

\end{document}